\documentclass[a4paper,10.5pt]{elsarticle}

\usepackage{CJKutf8}

\usepackage{geometry}
\usepackage{makecell}
\usepackage{extarrows}
\usepackage{subfigure}
\usepackage{mathrsfs}
\usepackage{lineno}
\usepackage{ulem}
\usepackage{graphicx}
\usepackage{subfigure}
\usepackage{fancyhdr}
\usepackage{amssymb}
\usepackage{multicol}
\usepackage{amsmath}
\usepackage[figuresright]{rotating}

\usepackage{hyperref}
\usepackage{cleveref}
\usepackage{multirow}

\usepackage{inputenc}

\begin{document}
\begin{frontmatter}

\title{A self-adaptive and robust fission clustering algorithm via heat diffusion and maximal turning angle}

\author{Yu Han$^{1}$, Shizhan Lu$^2$, Haiyan Xu$^{1*}$}

\address{\small  $^1$College of Economics and Management, Nanjing University of Aeronautics and Astronautics, Nanjing 211106, P.R. China  \\

\small  $^2$School of Economics and Management, Nanjing University of Science and Technology, Xiaolingwei 200, Nanjing 210094, P.R. China \\


}


\begin{abstract}
Cluster analysis, which focuses on the grouping and categorization of similar elements, is widely used in various fields of research. A novel and fast clustering algorithm, fission clustering algorithm, is proposed in recent year. In this article, we propose a robust fission clustering (RFC) algorithm and  a self-adaptive noise identification method. The RFC and the self-adaptive noise identification method are combine to propose a self-adaptive robust fission clustering (SARFC) algorithm. Several frequently-used datasets were applied to test the performance of the proposed clustering approach and to compare the results with those of other algorithms. The comprehensive comparisons indicate that the proposed method has advantages over other common methods.

\end{abstract}
\end{frontmatter}

\begin{multicols}{2}
\section{Introduction}

Clustering analysis,  as a useful technique in data mining  focusing on grouping and categorizing similar elements in a data set, has been  widely applied in diverse research areas,  ranging from climate predictions\cite{KCP}, gene expression\cite{WTX},  bioinformatics\cite{PEA}, finance and economics\cite{HJD, LG} to neuroscience\cite{GS, AHK}, to list a few.
Benefited  from the modern automatic monitoring technologies, people have gathered more and more  data. However,  with poor information of this data we   rarely     exactly know in advance  the number of clusters  before any
analysis.  Therefore,     a clustering method that does not need setting an input parameter or parameters is preferred, which is what we aim for in this article.

There have been various different clustering methods in the literature.
In general, they can be classified into 6 categories, that is
density-based
(c.f., DP\cite{RA}, DP-HD\cite{MR}, DBSCAN\cite{EM}, NQ-DBSCAN\cite{CHEN},  CSSub\cite{ZHY} and GDPC\cite{JJH}),
grid-based
(c.f., CLIQUE\cite{AGR}, Gridwave\cite{DEC} and WaveCluster\cite{SG}),
model-based
(c.f., Gaussian parsimonious\cite{MUK}, Gaussian mixture models\cite{OHA}  and Latent tree models\cite{CT}),
partitioning (c.f., K-means\cite{MJ, ZHR, ZHAR}, K-partitioning\cite{CDW} and TLBO\cite{LK}),
graph-based (SEGC\cite{WAJ}, ProClust\cite{PPI} and MCSSGC\cite{VIE}),
and hierarchical (c.f., BIRCH\cite{ZHT}  and CHAMELEON\cite{GAK}) approaches.

 Lu, Cheng and Mehmood\cite{LUSZ} proposed a  novel and interesting  clustering algorithm which can cluster data without any parameters, fission clustering (FC) algorithm. FC has several excellent features to compare with some existing algorithms: (F1),  FC algorithm has stronger coping ability for the extreme challenge data sets to compare with many of current clustering algorithms; (F2), the time complexity of FC is  superior to many of current existing algorithms; (F3), FC can utilize computing group to reduce the difficulty of data processing when it faces  big data sets in the future. More detailed descriptions are shown in section 2.

\begin{figure*}[hbtp]
\centering
\includegraphics[scale=0.46]{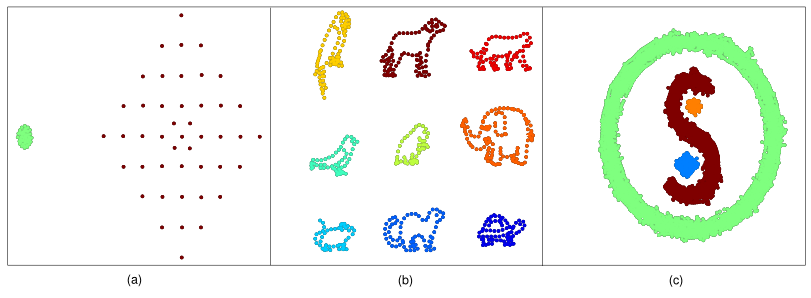}
\caption{Distribution shapes of data sets Imbalance, Animals, Ring}
\label{}
\end{figure*}

 However, FC only can apply in idealized data sets, i.e., the border points of $i$th cluster are far away from the border points of $j$th cluster. When  FC deal with the  general data sets, it needs  two parameters  (such as FC-KNN\cite{LUSZ}). The algorithms which need parameters are difficult applied to the data sets that obtain by automatic monitoring  devices. In the following,  a self-adaptive noise identification method  and a robust fission clustering algorithm are proposed, and then they are combined together to propose a self-adaptive and robust fission clustering (SARFC) algorithm. SARFC retains the performance of FC and it can deal with all the density-based data sets without any parameters.

The remainder of this article is organized as follows: Section 2 presents the relevant   knowledge of the fission clustering algorithm, heat diffusion density factor and self-adaptive noise identification method of linear regression. Section 3 proposes a robust fission clustering algorithm, a self-adaptive noise identification method of maximal turning angle, and a self-adaptive and robust fission clustering (SARFC) algorithm.  Section 4  demonstrates our algorithms by some numerical experiments of both simulated and real data sets and image segmentation data through comparison with some state-of-the-art clustering algorithms. Section 5 makes a conclusion.

\section{Literature  Research}

\subsection{Fission clustering (FC)}

FC that proposed by Lu, Cheng and Mehmood\cite{LUSZ} has several excellent feature, such as the following F1, F2 and F3.

{\bf F1.}  To compare with many of current clustering algorithms, FC algorithm has stronger coping ability for the extreme challenge data sets. As shown in Figure 1, there are three data sets that are constructed for increasing the clustering difficulty. The Imbalance data set magnifies the differences of two clusters' densities to increase the  clustering difficulty. Many of current density-based algorithms cannot discover the cluster with small density and identify Imbalance as one category data. Many of  current grid-based   algorithms classify Imbalance as more than two clusters data. The points in the cluster with small density are divided in different grid cells and cannot be combined, because the huge differences of two clusters' densities are out of the identification ability for grid-based algorithms with the same size of the grid. The Animals data set has several profiles of animals, it is constructed for increasing the difficulty of pattern recognition. Many of  current model-based algorithms are difficult to learn the correct pattern in Animals data set and cannot obtain the correct clustering result. As shown in Figure 1 (c), there is no single point can be considered as the geometrical centroid of the annulus in the Ring data set, the points of the "S" cluster interleave through other two cluster. Many of current partitioning  algorithms cannot find out the central point for every cluster in data set Ring, and obtain the wrong clustering result. As shown in Figure 1, FC algorithm can correctly cluster these three data sets with extreme challenge.

{\bf F2.} As shown in Figure 2, the processes above clustering result in Figure 2 are the clustering processes of FC algorithm, they  are similar to atomic fissions.  FC algorithm can obtain $k$ clusters after $k-1$ times subset partitions, its time complexity is $\mathcal{O}(k)$. The processes under clustering result in Figure 2 are similar to atomic fusions. The algorithms used this kind of clustering processes need to combine data point by point, their time complexities are at least $\mathcal{O}(n)$. FC algorithm has a big advantage in time complexity.

\begin{figure*}[hbtp]
\centering
\includegraphics[scale=0.46]{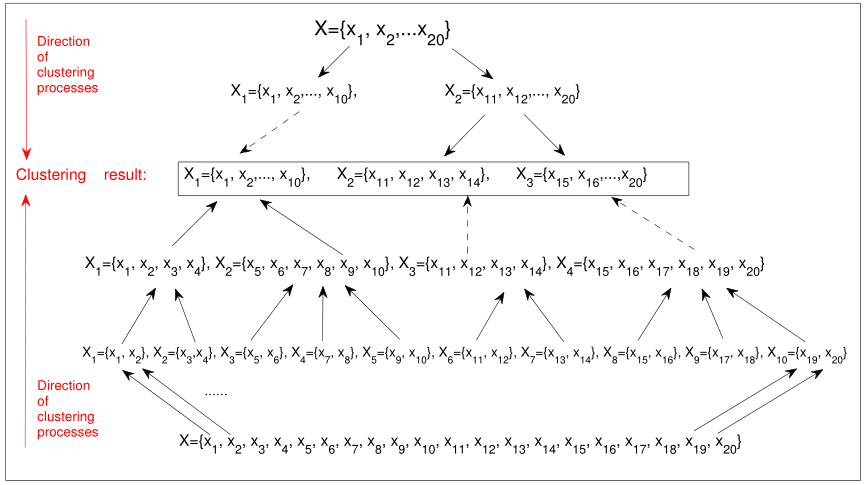}
\caption{The clustering processes of two different directions}
\label{}
\end{figure*}

{\bf F3.}  As shown in Figure 2, FC clusters data via splitting data sets. When FC faces the sets with billions of samples, it can use computing group to reduce difficulty of data processing. FC divides data set into subsets and then sends the subsets to the members of computing group to deal with subsets respectively, such as, computer 1 deals with the subset $C_1$, computer 2 deals with the subset $C_2$, and so on. The algorithms which cluster data with the pattern of atomic fusion cannot use computing group to optimize the procedures of data processing for big data sets.

Although FC has many advantages to compare with other algorithms, but it also has two disadvantages. (i) If a category in the data set has more than two points which keep away from other points in this category, FC will consider these outliers as another category, then one category is considered as two categories (as shown in Figure 4). (ii) If FC deals with the nonidealized data sets, it needs two parameters $k,t$\cite{LUSZ}. The algorithms needed parameters are not befitting for the automatic monitoring data. The section 3 of this article presents an improved FC algorithm to solve the issues (i) and (ii).

\subsection{Density factor of heat diffusion}

Nonparametric density estimation is an important tool for
statistical analysis of data. It is used to evaluate skewness, multi-modality, summarizing Bayesian posteriors, discriminant analysis,
and classification\cite{BZI,LELE}. Nonparametric approaches are more
flexible for modelling of datasets and are not affected by specification bias\cite{LELE}, in contrast to the classical approach\cite{BZI}. Kernel
density estimation (KDE) is the most commonly used nonparametric density estimation method\cite{BZI}. Mehmood et al\cite{MR} utilized KDE to proposed a method to calculate the density of every point in data set. Mehmood's method can estimate the density based on the nature of the data set without any parameter, it overcomes the aforementioned issues that method is self-adaptive for various data sets obtained by automatic monitoring devices.

KDE demonstrates a prominent  performance for mining the local dense families of sample set.
The KDE for an identical and independent data set  $X=\{x_1,x_2,\cdots,x_n\}$ drawn with an unknown probability density function (PDF) is

\begin{equation}
\hat{f}_h(x;h)=\frac{1}{n}\sum\limits_{j=1}^nK_h(x-x_j),
\end{equation}

where $K_h(x-x_j)$ is the kernel function and the Gaussian kernel $K(x,x_j;h)=\frac{1}{\sqrt{2\pi h}}e^{-\frac{(x-x_j)^2}{2h}}$ is normally used to estimate the density. The interpretation of KDE via heat diffusion derives from the concept of the Weiner process, interpreted as the probability distribution function at time $t$ of this process.

\begin{equation}
\hat{f}_t(x;t)=\frac{1}{2}\sum\limits_{j=1}^n\frac{1}{\sqrt{2\pi t}}e^{-\frac{(x-x_j)^2}{2t}}.
\end{equation}

Equation (2) is an iterative process that is similar to equation (1) with  bandwidth $h$, it has an approximate solution as below\cite{MR},

\begin{equation}
\hat{f}(x;t)\approx\sum\limits_{k=0}^{n-1} a_k e^{-k^2\pi^2t/2}\cos(k\pi x),
\end{equation}

where $n$ is a large positive integer, and $a_k=\frac{1}{n}\sum_{i=1}^n \cos(k\pi x_i),\ k=0,1,2,\cdots,n-1$.

Botev et al\cite{BZI} applied the improved Sheather-Jones algorithm to obtain a unique solution of the nonlinear equation. This has achieved by finding adaptively the optimal bandwidth $t$ for KDE,

\begin{equation}
t=\xi\gamma^{[l]}(t).
\end{equation}

The detailed description of the improved Sheather-Jones algorithm is shown in\cite{BZI}.

The role of Equation (3) is applied to obtain an indicator (denoted as $\rho_i$) for every object in the clustering sample set. A sample $x_i$ which has a larger value of $\rho_i$ is surrounded by many samples in the data set.

\subsection{Self-adaptive method of noise identification}

Lu and Cheng et al\cite{LUSZg} created an interesting and effective method to distinguish the border objects and dense area objects. They  researched the density variations of two kinds of objects, and then utilized smooth sequence of density sequence to make linear regression for noise identification (border objects identification), as shown in follows.

\begin{figure*}[hbtp]
\centering
\includegraphics[scale=0.46]{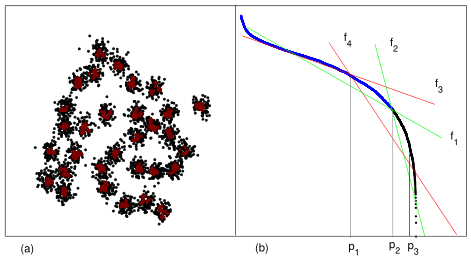}
	\caption{The distinctions of border objects and dense area objects showed in $V$. {\small((a) shows the border objects (black points) and dense area objects (colourful points) in data set. (b) shows two distinctly different distribution trends of $\{v_i\}=V$ which can be used to distinguish the border objects and dense area objects.)}}
\end{figure*}

They utilized grey correlation analysis to obtain the density sequence $\rho=\{\rho_i\}_n$ for data set $X$, and then smoothed the density sequence $\rho$ to sequence $V$. Let $\rho'=\{\rho_i': 1\leqslant i\leqslant n\}$ be the descending sequence of $\rho=\{\rho_j: 1\leqslant j\leqslant n\}$, and $V=\{v_i: 5\leqslant i\leqslant n\}$ be the mean smoothing sequence of $\rho'$, where $v_i=(\rho_i'+\rho_{i-1}'+\rho_{i-2}'+\rho_{i-3}'+\rho_{i-4}')/5$.

As shown in Fig. 3 (b), the elements of $V$ have two distinctly different distribution trends. The elements in $V$ are divided into two parts  $V_{p}^-=\{v_i:5\leqslant i\leqslant p\}$ and $V_{p}^+=\{v_i:p<i\leqslant n\}$. Then $(I_{p}^-,V_{p}^-)$ and $(I_{p}^+,V_{p}^+)$ are used to obtain two models of linear regressions, where $I_{p}^-=\{i:5\leqslant i\leqslant p\}$ and $I_{p}^+=\{i:p<i\leqslant n\}$. Let $f_1$ and $f_2$ be
the regression equations for $(I_{p}^-,V_{p}^-)$ and $(I_{p}^+,V_{p}^+)$, respectively. Then define
\begin{equation}
\left\{
\begin{aligned}
e_1(i)=f_1(i)-v_i, \  5\leqslant i\leqslant p, \\
e_2(i)=f_2(i)-v_i,\  p<i\leqslant n,
\end{aligned}
\right.
\end{equation}
and $R_p=\sum\limits_{5\leqslant i\leqslant p} |e_1(i)|+\sum\limits_{p<i\leqslant n}|e_2(i)|$, $R_p$ is  the  sum of absolute residuals (SoAR),  depending on $p$.

As illustrated
in Fig. 3, to pick $p_1$, $p_2$ and $p_3$ in different positions for linear regressions and obtain three SoARs $R_{p_1}$, $R_{p_2}$ and $R_{p_3}$,
where, as shown, $R_{p_2}<R_{p_1}$ and $R_{p_2}<R_{p_3}$.

The different positions $p_i$ can be used to generate
a sequence of SoARs, $\mathfrak{R}=\{R_i:5+5\leqslant i\leqslant n-5\}$ (5 boundary points left for boundary regression). If for some $p$, $R_p=min(\mathfrak{R})$, then  the object $x_j$ is considered as a member of the dense subset for the case $\rho_j\geqslant \rho'_p$. Then, an idealized dense subset is obtained for FC algorithm.

\section{Proposed methods}

\subsection{Robust fission clustering algorithm}

Lu, Cheng and  Mehmood\cite{LUSZ} utilized the following Definition 1 to divide set into two subsets, utilized Theorem 1 to stop dividing the subsets when $k$ subsets was obtained. However, FC algorithm is not enough robust. As shown in Figure 4 (a), $x_1$ and $x_2$ are two points that tinily leave the left cluster  in (a), FC considers $x_1$ and $x_2$ as an individual cluster. $x_3$, $x_4$ and $x_5$ are also considered as an individual cluster in (b). These points that  tinily leave cluster cannot distinctly consider as individual cluster. We will propose a robust fission clustering (RFC) algorithm in the following.

{\bf  Definition 1}\cite{LUSZ}. $f:X\times X\rightarrow R$ is a distance (similarity) function, where $X$ is a sample set, and $R$ is the real number set. For all $x_k\in X$, if  $f(x_0,x_k)\not\in(f(x_0,x_i),f(x_0,x_j))$ (or $f(x_0,x_k)\not\in(f(x_0,x_j),f(x_0,x_i))$), we call $|f(x_0,x_i)-f(x_0,x_j)|$ a crack of $(X,f)$, where $x_0,x_i,x_j\in X$.

{\bf Theorem 1}\cite{LUSZ}. If the distance (similarity) function $f$ satisfies triangle inequality and $C\subset X$ has a $d_0(C)$-road, then $MC(C)\leqslant d_0(C)$, where $MC(C)$ is the $MC$ of $(C,f)$.

\begin{figure*}
\centering
\includegraphics[scale=0.46]{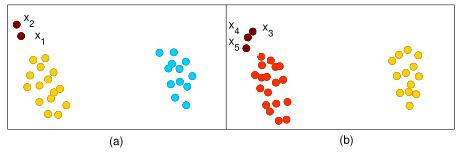}
	\caption{Some points that tinily leave their clusters}
	\label{}
\end{figure*}

To develop the RFC, some symbols and Definition 2 are necessary.

$\hat{x_i}^{(r)}$ is denoted as the $r$th near neighbor of $x_i$. Let $d_0^{(m)}(C)=max\{f(x_i,\hat{x_i}^{(m)}):x_i\in C\subset X\}$, and $d_0^{(m)}=max\{f(x_i,\hat{x_i}^{(m)}):x_i\in X\}$. $S(X)$ is denoted as the similarity matrix with respect to similarity function $f$. $S_1(X)$ is obtained by sorting every row of $S(X)$ with ascending. The $i$th column of $S_1(X)$ is subtracted from the $(i+1)$th column of $S_1(X)$ to obtain the $i$th column of the matrix $S_2(X)$. $MC^{(r)}(X)=max\{S_2^{(r)}(X)(i,j):i=1,2,\cdots,n, j=1,2,\cdots,n-r\}$.

{\bf  Definition 2.} If there is a connected component in $C$ such that there is a path from $x$ to $y$ for all $x,y\in C$ and the distance of random $r$ adjacent points $x_{p+1},x_{p+2},\cdots,x_{p+r}$ in this path is  less than or equal $d_0^{(r)}(C)$ (i.e. $f(x_p,x_{p+r})\leqslant d_0^{(r)}(C)$). This connected component is called $d_0^{(r)}(C)$-connected component (briefly $d_0^{(r)}(C)$-CC).

{\bf Theorem 2.} If the similarity function $f$ satisfies triangle inequality and $C\subset X$ has a $d_0^{(r)}(C)$-CC, then $MC^{(r)}(C)\leqslant d_0^{(r)}(C)$.

{\textit{Proof.}} If there are $i$ and $p$ such that $S_2^{(r)}(C)(i,p)=|f(x_i,x_{p+r})-f(x_i,x_p)|>d_0^{(r)}(C)$ is hold. Since $f$ satisfies triangle inequality and $C$ has a $d_0^{(r)}(C)$-CC, then $x_{p+r}$ is not the $r$th near neighbor of $x_p$. There must exists a $x_t$ such that $x_{p+r}$ is the $r$th near neighbor of $x_t$ in a path of $d_0^{(r)}(C)$-CC.

The $x_t$ can be described in two cases: (i) $f(x_i,x_t)\in(f(x_i,x_{p+j}),f(x_i,x_{p+j+1}))$; (ii) $f(x_i,x_t)=f(x_i,x_{p+j})$, where $0\leqslant j\leqslant r-1$.

For the case (i). $MC^{(r)}(C)$ is constituted by $r$ adjacent $MC(C)$. By Definition 1, there is no $x_t\in C$ such that $f(x_i,x_t)\in(f(x_i,x_{p+j}),f(x_i,x_{p+j+1}))$. A contradiction.

For case (ii). If $f(x_i,x_t)=f(x_i,x_{p+j})$, then $|f(x_i,x_{p+r})-f(x_i,x_p)|=S_1(C)(i,p+r+1)-S_1(C)(i,p)$. Although the mappings $f(x_i,x_t)$ and $f(x_i,x_{p+j})$ are coincident, but there are $r+1$ mappings in $(f(x_i,x_p),f(x_i,x_{p+r}))$, it is a contradiction with the construction of $S_2(C)$.

\begin{figure*}
\centering
\includegraphics[scale=0.46]{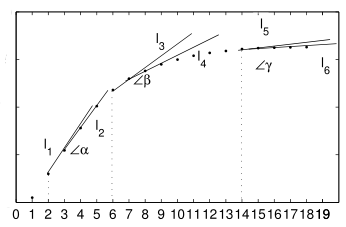}
	\caption{The turning angles in curve $V$}
	\label{}
\end{figure*}

Hence, $S_2^{(r)}(C)(i,p)\leqslant d_0^{(r)}(C)$ is hold for all $i$ and $p$, i.e., $MC^{(r)}(C)\leqslant d_0^{(r)}(C)$.
$\blacksquare$

Note that, FC in\cite{LUSZ} is only a specific case of our RFC. When $r=1$,  RFC and FC are same.

To look back the issue that describes in Figure 4. When $r=2$, RFC can clusters the data set in Figure 4 (a) as two categories, not three categories; but also
clusters the data set in Figure 4 (b) as three categories. When $r=3$, RFC can clusters the data sets in Figure 4 (a) and (b) as two categories. The distinctions of RFC and FC are described as following.
\begin{tabular}{p{106pt}| p{111pt}}
\hline
FC &  RFC \\
\hline
{\small Stop dividing subsets with the condition $d_0=max\{f(x_i,\hat{x_i}):x_i\in X\}$ } &{\small  Stop dividing subsets with the condition  $d_0^{(r)}=max\{f(x_i,\hat{x_i}^{(r)}):x_i\in X\}$ } \\
\hline
{\small  The construction of $MC(C)$, $S_2(C)(:,k)=\newline S_1(C)(:,k+1)-S_1(C)(:,k)$ } & {\small  The construction of $MC^{(r)}(C)$, $S_2^{(r)}(C)(:,k)=\newline S_1(C)(:,k+r)-S_1(C)(:,k)$ }\\
\hline
{\small  Allow $r$ ($r\geqslant2$) points as an autocephalous category }& {\small Avoid $m$ ($m\leqslant r$) points to be an autocephalous category }  \\
\hline
\end{tabular}\\

\rule{7.6cm}{0.5mm}

{\bf Algorithm 1:} RFC algorithm.

\rule{7.6cm}{0.25mm}

{\bf Input:} Similarity (distance) matrix $S(X)$.

{\bf Output:} Clusters of $X$.

1. If $|X|\leqslant1000$, $r\leftarrow1$; if $1000<|X|\leqslant2000$, $r\leftarrow2$; if $|X|>2000$, $r\leftarrow3$.

2. $d_0^{(r)}\leftarrow max\{f(x_i,\hat{x_i}^{(r)}):x_i\in X\}$.

3. $C_1\leftarrow X$ (initial value).

4. {\bf While}  There is a subset $C_i$ such that $MC^{(r)}(C_i)>d_0^{(r)}$ {\bf do}

5.  {\bf repeat}

6.  Pick the subset $C_i$ if $MC^{(r)}(C_i)>d_0^{(r)}$.

7. Sort every row of $S(C_i)$ to obtain $S_1(C_i)$.

8. $S_2^{(r)}(C_i)(:,k)\leftarrow S_1(C_i)(:,k+r)-S_1(C_i)(:,k)$, $k=1,2,\cdots,n-r$.

9. $MC^{(r)}(C_i)\leftarrow max\{S_2^{(r)}(C_i)(k,j):k=1,2,\cdots,n,\ j=1,2,\cdots,n-r\}$.

10. Find out $x_i,x_j$ and $x_k$ which generate the $MC^{(r)}(C_i)$ ($|f(x_i,x_j)-f(x_i,x_k)|=MC^{(r)}(C_i)$).

11. If $f(x_i,x_t)\leqslant min\{f(x_i,x_j),f(x_i,x_k)\}$ then $x_t\in C_{count}$; otherwise, $x_t\in C_{count+1}$.

12.  {\bf until} $max\{MC^{(r)}(C_i)\}\leqslant d_0^{(r)}$.

13.  {\bf end while}

\rule{7.6cm}{0.25mm}\\

The details of the RFC algorithm are as shown in follows, where $S_k(C)(:,i)$ is the $i$th column of $S_k(C)$.

\subsection{An improved method for self-adaptive noise identification}

$V$ is described in subsection 2.3. As shown in Figure 5, $l_1$ is the straight line which passes two points $(2,v_2)$ and  $(3,v_3)$, $l_2$ is the straight line which passes two points $(3,v_3)$ and  $(4,v_4)$. The intersection angle of $l_1$ and $l_2$ is $\angle\alpha$. $\angle\alpha$ is a turning angles which describes the veer of $v_2$, $v_3$ and $v_4$. As shown in Figure 5, the points $v_i$ of $V$ have two distinctly different distribution trends. The position which has a maximal turning angle can be used to distinguish the points of two different distribution trends. We can applied the maximal turning angle position to distinguish the dense area points and border points.

If the sequence $V=\{v_i\}$ is smooth enough, the turning angles will be very small (as shown in Figure 5). It is ensure that the turning angles are fall in interval $[0,\pi/2]$.  When $0\leqslant\angle\alpha_i\leqslant\pi/2$, $\angle\alpha_i$ is positively related with $\tan\angle\alpha_i$. The self-adaptive algorithm that utilize maximal turning angle to recognize noise is shown in follows.

\begin{figure*}
\centering
\includegraphics[scale=0.46]{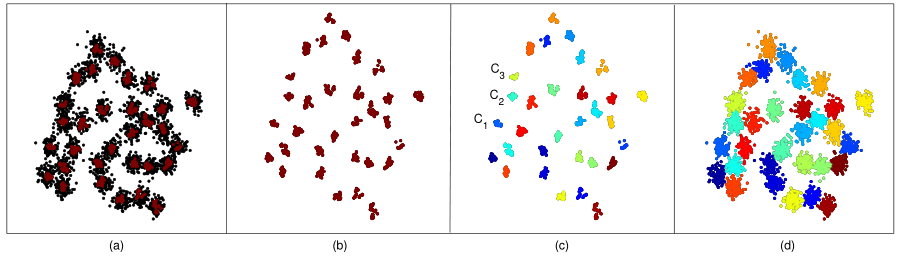}
	\caption{The main processes of the SARFC}
	\label{}
\end{figure*}

\rule{7.6cm}{0.5mm}

{\bf Algorithm 2:} Self-adaptive noise identification algorithm.

\rule{7.6cm}{0.25mm}

{\bf Input:} Similarity (distance) matrix $S(X)$.

{\bf Output:} Dense subset $C$.

1. Apply  equation (3)  to obtain $\rho_i$ for every object.

2. Sort the sequence $\rho=\{\rho_i\}$ in descending order to obtain sequence $\rho'=\{\rho'_i\}$.

3. Smooth the sequence $\rho'$ to obtain sequence $V=\{v_i\}$, where $v_i=(\rho_i'+\rho_{i-1}'+\rho_{i-2}'+\rho_{i-3}'+\rho_{i-4}')/5$.

4. Utilize the method in Section 2.3 to obtain a position $p_r$.

5. Obtain sequence $H=\{h_i\}$, where $h_i=v_{i+1}-v_i$.

6. Obtain sequence $K=\{k_i\}$, where $k_i=h_i/\Delta T$, $\Delta T$ is the length of horizontal axis between $\rho_{i+1}$ and $\rho_i$,  $\Delta T=1$ in this article.

7. Calculate tangent sequence $\{\tan\alpha_i\}$ for turning angles, where $\tan\alpha_i=|(k_i-k_{i-1})/(1+k_ik_{i-1})|$.

8. $p_{max}\leftarrow 1$.

9. {\bf While}  $p_{max}<p_r$ {\bf do}

10. {\bf repeat}

11. $\tan\alpha_{p_{max}}\leftarrow 0$, and update tangent sequence $\{\tan\alpha_i\}$.

12. Find out the maximum of $\{\tan\alpha_i\}$ and the position $p_{max}$ of the maximum.

13.  {\bf until} $p_{max}>p_r$.

14.  {\bf end while}

15. If $\rho_j\geqslant\rho'_{p_{max}}$,  the object $x_j$ is considered as a member of the dense subset, then the dense subset $C$ is obtained.

\rule{7.6cm}{0.25mm}\\

\subsection{A self-adaptive and robust fission clustering (SARFC) algorithm}

The main steps of the SARFC algorithm are as shown in follows.

\rule{7.6cm}{0.5mm}

{\bf Algorithm 3:} SARFC algorithm.

\rule{7.6cm}{0.25mm}

{\bf Input:} Similarity (distance) matrix $S(X)$.

{\bf Output:} Clusters of $X$.

1. Apply Algorithm 2 to obtain a dense subset $C$.

2. Cluster the subset $C$ by using Algorithm 1.

3. {\bf While} $X-C\neq\emptyset$, {\bf do}

4.  {\bf repeat}

5. Assign $x'_j$ into the cluster which contains the object $x'_i$, if $f(x'_i,x'_j)=min\{f(x_i,x_j):x_i\in C, x_j\in X-C\}$.

6. Update $C$ and $X-C$, $C\leftarrow C\cup\{x'_j\}$, $X-C\leftarrow X-C-\{x'_j\}$.

7. {\bf until} $X-C=\emptyset$.

8. {\bf end while}.

\rule{7.6cm}{0.25mm}\\

\begin{figure*}
\centering
\includegraphics[scale=0.46]{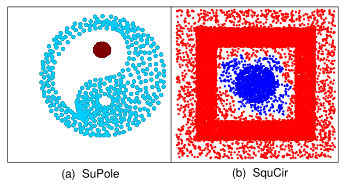}
	\caption{Distribution shapes of data sets SuPole and SquCir}
	\label{}
\end{figure*}

The main processes of the SARFC can be displayed as Figure 6. Figure 6 (a) to (b) shows the processes that Algorithm 2 obtains a dense subset $C$ from the data set $X$. Figure 6 (b) to (c) shows the processes that Algorithm 1 clusters the dense subset $C$. Figure 6 (c) to (d) shows the processes that assign the object in $X-C$ to its nearest category.

Equation (3) can be solved using fast Fourier transform and takes  $\mathscr{O}(nlogn)$ operations.  The methods of Section 2.3 and 3.2 obtain $p_r$ and $p_{max}$ take $\mathscr{O}(n)$ operations, respectively. The time complexity of Algorithm 2 is $\mathscr{O}(nlogn)$. Algorithm 1 splits the set $X$ (or dense subset $C$) into subsets $C_1,C_2,\cdots,C_k$. Since $|C_i|\ll |X|$, data processing will become faster and faster, accompanied by the dividing courses of subsets.  $k$ clusters are obtained after $k-1$ times of dividing subsets, then, the time complexity of Algorithm 1 is $\mathscr{O}(1)$. Moreover, if $|C|=m$, then $|X-C|=n-m$, and the time complexity of assigning border points is hence $\mathscr{O}(n-m)$. The time complexity of the whole SARFC algorithm is $\mathscr{O}(nlogn)$.

\section{Experiments}

Some classical or state-of-the-art density-based algorithms will be selected to compare with the proposed density-based algorithm: SARFC algorithm. Such as, DP\cite{RA}, ADPC\cite{TLI}, RNN-DBSCAN\cite{BACK}, MC-FNGLVQ\cite{MASU} and UNIC\cite{LNRO}. The classical density-based algorithm DP\cite{RA}, and DP's state-of-the-art improved algorithm ADPC\cite{TLI}. RNN-DBSCAN\cite{BACK} is the self-adaptive improved algorithm of the classical density-based algorithm DBSCAN\cite{EM} (not self-adaptive). Moreover, an up to date intelligence algorithm MC-FNGLVQ\cite{MASU} and a base on statistical algorithm UNIC\cite{LNRO} are selected to be the added comparison algorithm.

\subsection{Experiments on synthetic and real data sets}

\subsubsection{Descriptions of data sets}

In this part, some frequently-used data sets obtained from different clustering references are applied to test the algorithms. Such as R15\cite{VCJ}, D31\cite{VCJ},  Aggregation (Agg)\cite{GA}, A1\cite{IK} and S1\cite{FP}  etc. And then two datasets, SuPole and SquCir (Figure 7), are constructed for the supplementary tests. All the simulation data are points of two-dimensional Euclidean space.

Three real-world data sets Iris\footnote{http://archive.ics.uci.edu/ml/datasets.php}\cite{FRA, HF}, Seeds$^1$\cite{CM} and Wine $^1$\cite{ZHP} are applied to make data experiments for algorithms comparisons. Data set Iris contains three kinds of irises: Setosa, Versicolour and Virginica, each kind has 50 samples. Data set Seed is obtained  from three places: Kama,  Rosa and Canadian, 70 seeds from each place. Data set Wine are the results of a chemical analysis of wines grown in the same region in Italy but derived from three different cultivars. The description of the data sets are shown in Table 1.

\begin{tabular}{p{56pt}|p{43pt} p{43pt} p{43pt}}
\hline
{\footnotesize Data}&{\footnotesize Instances} &{\footnotesize Features} & {\footnotesize Clusters}    \\
\hline
R15 & 600 &2  & 15 \\
D31 &3100  & 2 & 31 \\
Agg & 788 & 2 & 7 \\
A1 & 3000 & 2 & 20 \\
S1 & 5000 & 2 & 15 \\
SuPole & 513 & 2 & 2 \\
SquCir & 50000 & 2  & 2 \\
Iris& 150 & 4 & 3 \\
Seeds& 210 & 7 & 3 \\
Wine& 178 & 13 & 3  \\
\hline
\end{tabular}
{\centering\footnotesize Table 1: The simple description of data sets}

\subsubsection{Presentation of results}

Table 2 presents the number of clusters estimated by various methods. Table 3 shows the clustering results when compared with other various methods.

To evaluate and compare the performance of the clustering methods, we apply the evaluation metrics: Accuracy (Acc), F$_1$-Score (F$_1$),  Adjusted Rand Index (ARI)\cite{DUL} and Normalized Mutual Information (NMI)\cite{ABB} in our experiments to do a comprehensive evaluation.   The higher
the value, the better the clustering performance for all these measures.
Compared with the best results of other algorithms,   our method has  relative advantages of 7.4\%, 5.5\%, 11.5\% and 7.2\% (Table 3) with respect to Accuracy, F$_1$-Score, ARI and NMI for the Iris dataset, respectively.

In summary, our method achieves better results with respect to the estimation of cluster number, Acc, F$_1$, ARI and NMI, compared comprehensively with other methods.

\begin{tabular}{p{26pt}|p{15pt} p{18pt}p{28pt}p{26pt}p{20pt}p{20pt}}
\hline
{\footnotesize Data}&{\footnotesize DP} &{\footnotesize ADPC} &{\footnotesize RNN-DBSCAN} &{\footnotesize MC-FNGLVQ} & {\footnotesize UNIC}& {\footnotesize SARFC} \\
\hline
R15 & {\bf 15} &{\bf 15} &{\bf 15} & 9 &{\bf 15}&{\bf 15}\\
D31 & {\bf 31} &{\bf 31}  & {\bf 31}  &23 & {\bf 31} &{\bf 31} \\
Agg & {\bf 7}  &{\bf 7}   & {\bf 7}  &8  &  {\bf 7}&{\bf 7}\\
A1 & {\bf 20}  & {\bf 20}  &  18 & 18 & 17 &{\bf 20} \\
S1 & {\bf 15}  & {\bf 15}  & 8   &11  &  9 & {\bf 15}\\
SuPole & 5  & 3 &  15  &  3& 5 &{\bf 2} \\
SquCir & 9  & {\bf 2}    &  {\bf 2}  & 15  & 6 &{\bf 2} \\
Iris& 2 &  2 &  4  & 6 & 4 & {\bf 3}\\
Seeds& {\bf 3} & {\bf 3} &  4 & 5 & 4 &{\bf 3} \\
Wine&  7 & 6  & 5  & 6 & 5 & 4 \\
\hline
\end{tabular}
{\footnotesize Table 2: Number of clusters estimated by various algorithms}

\subsubsection{Comparisons}

We conduct the Friedman test with the post-hoc Nemenyi test ($\alpha$= 0.10)\cite{ZHY} to examine whether the difference between any two
clustering algorithms is significant in terms of their average ranks. The
difference between two algorithms is significant if the gap between their ranks is
larger than CD. There is a line between two algorithms if the rank gap between
them is smaller than CD. This test shows that SARFC is significantly better than the DP, RNN-DBSCAN, MC-FNGLVQ and UNIC. Except the SARFC, other algorithms are not significantly better than each other. The SARFC is the best performer of these algorithms, followed by the ADPC (as seen in Figure 8).

\begin{tabular}{p{20pt}|p{15pt}|p{12pt} p{16pt}p{28pt}p{28pt}p{15pt}p{15pt}}
\hline
{\scriptsize Data}& &{\scriptsize DP} &{\scriptsize ADPC} &{\scriptsize RNN-D\par BSCAN} &{\scriptsize MC-FN\par GLVQ} & {\scriptsize UNIC}& {\scriptsize SARFC} \\
\hline
{\footnotesize R15} &{\footnotesize  Acc} &99.2 &99.2 &98.5 &65.7&66.9&{\bf 99.3} \\
                   &{\footnotesize  $F_1$} &99.2 &99.2 &97.1 & 63.3 &64.3&{\bf 99.4}\\
                   & {\footnotesize ARI} &98.2 &98.2&98.4 & 65.2 &65.3&{\bf 98.6}\\
                   &{\footnotesize  NMI} &98.6 &98.6&98.2 & 61.3 &65.2&{\bf 98.9}\\
                   \hline
{\footnotesize D31} & {\footnotesize  Acc} &{\bf 96.8} &{\bf 96.8}  & 91.6 &53.2 & 68.0&{\bf 96.8}  \\
                  &{\footnotesize  $F_1$} &{\bf 96.8 }&{\bf 96.8} & 95.3 &49.9 & 68.6&{\bf 96.8 }\\
                  & {\footnotesize ARI} &{\bf 93.5}&{\bf 93.5} & 89.6 &50.9& 66.1 &{\bf 93.5} \\
                  & {\footnotesize  NMI}&{\bf 95.8} &{\bf 95.8} & 90.3 &52.1 & 70.0 &{\bf 95.8} \\
                  \hline
{\footnotesize Agg} & {\footnotesize  Acc} &99.9 &99.9 & 99.3  &87.3& 73.3 &{\bf 100 }\\
                  &{\footnotesize  $F_1$} &99.8 &99.8 & 98.9 &83.1 &71.7 &{\bf 100} \\
                  & {\footnotesize ARI} &99.8 &99.8 & 97.9  &90.1& 75.3 &{\bf 100 }\\
                  & {\footnotesize  NMI}&99.6 &99.6 & 99.9  &89.3& 73.6 &{\bf 100 }\\
                  \hline
{\footnotesize A1} & {\footnotesize  Acc} &94.3 &94.3 & 88.5  &85.2 & 78.0 &{\bf 97.2} \\
                  &{\footnotesize  $F_1$} &93.3 &93.3 &89.4 &83.7 & 80.5 &{\bf 97.2} \\
                  & {\footnotesize ARI} &92.1 &92.1 & 90.3  &87.2 & 77.9 &{\bf 94.6 }\\
                  & {\footnotesize  NMI}&93.1 &93.1 & 89.4 &89.0 & 81.0 &{\bf 96.2 }\\
                  \hline
{\footnotesize S1} & {\footnotesize  Acc} &92.6 &92.6  & 80.9 &89.3 & 85.5 &{\bf 99.3} \\
                  &{\footnotesize  $F_1$} &{\bf 93.3} &{\bf 93.3} & 82.2  &88.1 & 81.7 &{\bf 99.3 }\\
                  & {\footnotesize ARI} &89.2 &89.2  & 81.0  &86.9 & 82.6 &{\bf 98.6 }\\
                  & {\footnotesize  NMI}&94.5 &94.5 & 84.1  &89.6 & 83.0 &{\bf 99.0 }\\
                  \hline
{\footnotesize SuPole}& {\footnotesize  Acc} &47.6 &77.3  & 14.3 &71.7 & 37.1 &{\bf 100} \\
                  &{\footnotesize  $F_1$} &36.2 &77.8  & 14.1  &65.2 & 30.2 &{\bf 100} \\
                  & {\footnotesize ARI} &26.8 &27.1  & 29.6  &39.9 & 20.5&{\bf 100} \\
                  & {\footnotesize  NMI}&27.5 &31.4 &15.5  &28.8 & 20.9 &{\bf 100} \\
                  \hline
{\footnotesize SquCir} & {\footnotesize  Acc} &46.7 &58.2  & 57.3 &15.7 &26.1 &{\bf 98.6} \\
                  &{\footnotesize  $F_1$} &43.3 &55.6 & 56.2 &11.3 & 23.3 &{\bf 97.8} \\
                  & {\footnotesize ARI} &39.8&51.1 & 52.8  &9.8 & 20.8&{\bf 98.1} \\
                  & {\footnotesize  NMI}&37.2&52.8  & 53.6  &10.2& 21.1 &{\bf 98.3} \\
                  \hline
{\footnotesize Iris}& {\footnotesize  Acc} &66.7 &66.7 & 83.3  &69.3 &80.7 &{\bf 90.7}\\
                  &{\footnotesize  $F_1$} &57.1 &57.1  & 86.2  &72.1 & 84.5 &{\bf 91.7} \\
                  & {\footnotesize ARI} &56.8 &56.8 & 64.4 &33.6 & 60.9 &{\bf 75.9} \\
                  & {\footnotesize  NMI}&73.4 &73.4 & 73.4  &41.6 & 72.3 &{\bf 80.6 }\\
                  \hline
{\footnotesize Seeds}& {\footnotesize  Acc} &81.6 &81.6 & 85.7  &73.3 & 76.2 &{\bf 88.6} \\
                  &{\footnotesize  $F_1$} &83.1 &83.1  & 86.9 &76.5 & 78.5 &{\bf 89.0 }\\
                  & {\footnotesize ARI} &{\bf 80.1 }&{\bf 80.1} & 61.7 &37.8 & 43.0 &70.3 \\
                  & {\footnotesize  NMI}&{\bf 79.0} &{\bf 79.0 }& 62.6 &35.9 & 27.3 &69.8 \\
                  \hline
{\footnotesize Wine}& {\footnotesize  Acc} &39.9 &44.4  & 53.9  &55.1 & 56.2 &{\bf 61.2 }\\
                  &{\footnotesize  $F_1$} &53.9 &58.2 & 61.2  &{\bf 64.9 }& 63.8 &58.7 \\
                  & {\footnotesize ARI} &21.7 &23.9 & 21.6  &28.7 &23.3 &{\bf 31.6} \\
                  & {\footnotesize  NMI}&19.8 &23.4  & 14.0  &16.5 & 13.2 &{\bf 40.3} \\
\hline
\end{tabular}
{\footnotesize Table 3: The results' comparison  for different methods (\%)}

\begin{figure*}
\centering
\includegraphics[scale=0.46]{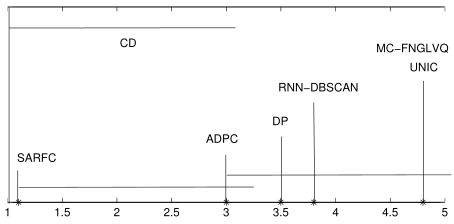}
	\caption{Critical difference (CD) diagram of the post-hoc Nemenyi test (at a significance level $\alpha$= 0.10)}
	\label{}
\end{figure*}

\section{Conclusion}

Benefit from the automatic monitoring technique, we will obtain more and more automatic monitoring data sets. Except the sample size and data dimensionality, we do not know to much information for this kind of data. Many current clustering algorithms need to input some parameters, but we do not know how to set the parameters for the automatic monitoring data. Hence, self-adaptive clustering algorithms are the future needs. In this article, a self-adaptive robust fission clustering (SARFC) algorithm is proposed. And then both the simulation and real datasets are applied to test the performance and effectiveness of the proposed method. Our proposed algorithm  is also compared with several frequently-used clustering algorithms. The experiments indicate that our method achieves better results, in terms of the evaluation metrics (Table 3) and the estimated number of clusters (Table 2), than the other methods under comparison.\\




\end{multicols}

{\bf REFERENCES}
\begin{thebibliography}{00}
\footnotesize

\bibitem{KCP}      CP Kappe,  M B\"{o}ttinger,  H Leitte. Analysis of Decadal Climate Predictions with User guided Hierarchical Ensemble Clustering.  Computer Graphics Forum, 2019, 38(3): 505-515.

\bibitem{WTX}    T Wang, J Zhang, K Huang. Generalized gene co-expression analysis via subspace clustering using low-rank representation.  BMC Bioinformatics, 2019,   20(7): 17-27.

\bibitem{PEA}    A Pessia, J Corander. Kpax3: Bayesian bi-clustering of large sequence datasets. Bioinformatics, 2018, 34(12):  2132-2133.

\bibitem{HJD} JD Hamilton.  A new approach to the economic analysis of nonstationary time series and
the business cycle.  Econometrica, 1989,  57(2): 357-384.

\bibitem{LG}  G Leibon, S Pauls, D Rockmore, R Savell.  Topological structures in the equities market
network.  Proceedings of the national academy of sciences of the united states of america, 2008,  105(52): 20589-20594.

\bibitem{GS} S Galbraith, JA Daniel, B. Vissel.  A study of clustered data and approaches to its analysis.
Journal of Neuroscience, 2010, 30(32):  10601-10608.

\bibitem{AHK}   HK Aljobouri, et al.  Clustering fMRI data with a robust unsupervised learning algorithm for neuroscience data mining. Journal of Neuroscience Methods, 2018,  299: 45-54.

 \bibitem{RA} A Rodriguez, A Laio. Clustering by fast search and find of density
peaks.  Science, 2014,   344(6191): 1492-1496.

\bibitem{MR} R Mehmood et al. Clustering by fast search and find of density peaks via heat diffusion.  Neurocomputing, 2016, 208: 210-217.

\bibitem{EM} M Ester, HP Kriegel, J Sander, X Xu.  A density-based algorithm
for discovering clusters in large spatial databases with noise.  Data Mining and Knowledge Discovery, 1996,  96(34):  226-231.

\bibitem{CHEN} Y Chen, S Tang, et al.  A fast clustering algorithm based on pruning unnecessary distance
computations in DBSCAN for high-dimensional data.  Pattern Recognition, 2018, 83: 375-387.



\bibitem{ZHY} Y Zhu, KM Ting, MJ Carman.  Grouping points by shared subspaces for effective subspace clustering.  Pattern Recognition, 2018,  83: 230-244.

\bibitem{JJH}  J Jiang, D Hao, et al. GDPC: Gravitation-based Density Peaks Clustering algorithm. Physica A, 2018, 502: 345-355.


\bibitem{AGR} R Agrawal, JE Gehrke, D Gunopulos, P Raghavan.  Automatic
subspace clustering of high dimensional data for data mining applications.
in Proc. ACM SIGMOD Int. Conf. Manage. Data (SIGMOD), Seattle, WA,
USA, 1998: 94-105.

\bibitem{DEC}  C Deng et al. Gridwave: a grid-based clustering algorithm for market transaction data based on spatial-temporal density-waves and synchronization. Multimedia Tools and Applications, 2018,  77(21): 1-15.

\bibitem{SG} G Sheikholeslami, S Chatterjee, A Zhang.  WaveCluster: a
wavelet-based clustering approach for spatial data in very
large data bases.  VLDB Journal, 2000,  8: 289-304.


\bibitem{MUK} K Murphy, TB Murphy. Gaussian parsimonious clustering models with covariates and a noise component. Advances in Data Analysis and Classification,  2019: 1-33.

\bibitem{OHA}  A O'Hagan et al. Investigation of Parameter Uncertainty in Clustering Using
a Gaussian Mixture Model Via Jackknife, Bootstrap and
Weighted Likelihood Bootstrap. Computational Statistics, 2019,  34(4): 1779-1813.

\bibitem{CT} T Chen, NL Zhang, T Liu, KM Poon, Y Wang.  Model-based
multidimensional clustering of categorical data.  Artificial Intelligence, 2012, 176(1):
  2246-2269.

\bibitem{MJ} J MacQueen. Some Methods for Classification and Analysis of Multivariate Observations. in Proceedings of the Fifth Berkeley Symposium
on Mathematical Statistics and Probability, LM Le Cam,
J Neyman, Eds. (Univ. California Press, Berkeley, CA), 1967,
 1: 281-297.

\bibitem{ZHR} R Zhang, X Li, et al. Deep Fuzzy K-Means with Adaptive Loss and Entropy Regularization. IEEE Transactions on Fuzzy Systems,   2019,  DOI:10.1109/TFUZZ.2019.2945232.

\bibitem{ZHAR} R Zhang et al. Joint Learning of Fuzzy k-Means and Nonnegative Spectral Clustering With Side Information. IEEE Transactions on Image Processing, 2019,   28(5): 2152-2162.

\bibitem{CDW}  DW Choi, CW Chung. A K-partitioning algorithm for clustering large-scale spatio-textual data. Information Systems, 2017, 64: 1-11.


\bibitem{LK}  K Lahari, MR Murty, SC Satapathy.  Partition based
clustering using genetic algorithm and teaching learning based optimization:
performance analysis.  Advances in Intelligent Systems and Computing, 2015, 338: 191-200.

\bibitem{WAJ} J Wang, W Zheng, Y Qian, J  Liang. A Seed Expansion Graph Clustering Method for Protein Complexes Detection in Protein Interaction Networks. Molecules, 2017, 22(12): 1-19.


\bibitem{PPI} P Pipenbacher, A Schliep, S Schneckener, et al.  ProClust: improved clustering of protein
sequences with an extended graph-based approach.  Bioinformatics, 2002,  18(2): 182-191.

\bibitem{VIE} VV Vu, HQ Du.   Graph-based Clustering with Background Knowledge.  SoICT 2017 Proceedings of the Eighth International Symposium on Information and Communication Technology,  NY, USA, 2017: 167-172.

\bibitem{ZHT} T Zhang, R Ramakrishnan, M Livny.  BIRCH: An efficient data
clustering method for very large databases.  in Proc. ACM SIGMOD
Int. Conf. Manage. Data (SIGMOD), Montreal, QC, Canada, 1996, 25(2): 103-114.


\bibitem{GAK} G Karypis, EH Han, V Kumar.  CHAMELEON: A hierarchical clustering algorithm
using dynamic modeling.  IEEE Computer, 1999, 32(8): 68-75.

\bibitem{LUSZ}  SZ Lu, LS Cheng, R Mehmood. Clustering by Using the Way of Atomic Fission. IEEE Access, 2020, 8: 70997-71007.


\bibitem{BZI} ZI Botev, JF Grotowski, DP Kroese.  Kernel density estimation via diffusion.
Annals of Statistics, 2010, 38(5): 2916-2957.

\bibitem{LELE}   LE Lehmann. Model specification: the views of Fisher and Neyman, and
later developments, Statistical Science, 1990, 5(2): 160-168.

\bibitem{LUSZg} SZ Lu, LS Cheng, ZD Lu, BA Khan. A Self-adaptive Grey DBSCAN Clustering.  2020. https://arxiv.org/abs/1912.11477.


\bibitem{TLI}  T  Liu, H  Li, X  Zhao.  Clustering by Search in Descending Order and
Automatic Find of Density Peaks. IEEE Access, 2019,   7: 133772-133780.




\bibitem{BACK}  A Bryant, K Cios. RNN-DBSCAN: A Density-Based Clustering Algorithm Using Reverse Nearest Neighbor Density Estimates[J]. IEEE Transactions on Knowledge and Data Engineering, 2018, 30(6): 1109-1121.


\bibitem{MASU}  MA  M\'{a}sum. Intelligent Clustering and Dynamic Incremental
Learning to Generate Multi-Codebook Fuzzy Neural
Network for Multi-Modal Data Classification[J]. Symmetry, 2020, 12(4): 679.

\bibitem{LNRO}   N Leopold, O Rose. UNIC: A fast nonparametric clustering[J]. Pattern Recognition, 2020, 100: 107-117.

\bibitem{VCJ} CJ Veenman,  MJT Reinders, E  Backer.  A maximum variance
cluster algorithm.  Ieee Transactions On Pattern Analysis And Machine Intelligence, 2002,   24(9): 1273-1280.


\bibitem{GA} A Gionis,  H Mannila,  P Tsaparas.  Clustering aggregation. ACM Transactions on Knowledge Discovery from Data,  2007,  1(1): 1-30.


\bibitem{IK} K Ismo, P Franti.  Dynamic local search for clustering with unknown number of clusters.  in: Proceedings of International Conference on
Pattern Recogn.,  2002,  2(16): 240-243.

\bibitem{FP} P Franti, O Virmajoki.  Iterative shrinking method for clustering problems.
Pattern Recognition, 2006,   39(5):  761-775.

\bibitem{FRA} RA Fisher.  The use of multiple measurements in taxonomic problems.
      Annual Eugenics, 1936,  7(2):  179-188.

\bibitem{HF} F Huang,  X Li, S Zhang, J Zhang.  Harmonious Genetic Clustering.  IEEE Transactions on Cybernetics, 2018,  48(1):  199-214.

\bibitem{CM} M Charytanowicz, J Niewczas, et al.  A Complete Gradient Clustering Algorithm for Features Analysis of X-ray Images.  in: Information Technologies in Biomedicine, Ewa Pietka, Jacek Kawa (eds.), Springer-Verlag, Berlin-Heidelberg, 2010, 15-24.


\bibitem{ZHP}  P Zhong, M Fukushima. Regularized nonsmooth Newton method for multi-class support vector machines. Optimization Methods and Software, 2007,  22(1): 225-236.

\bibitem{DUL}  L Du, Y Pan, X Luo.  Robust spectral clustering via matrix aggregation. IEEE Access, 2018,  6: 53661-53670.

\bibitem{ABB}  S  Abbasi, S  Nejatian, et al.   Clustering ensemble selection considering quality and diversity.  Artificial Intelligence Review, 2019,  52:  1311-1340.








\end{thebibliography}
\end{document}